\documentclass[times, 10pt,twocolumn]{article}
\usepackage{latex8}
\usepackage{times}
\usepackage{balance}
\usepackage{latexsym}
\usepackage{doc}
\usepackage{exscale}
\usepackage{fontenc}
\usepackage{graphicx}
\usepackage{epsfig}

\def\bx{{\mathbf x}}

\def\bk{{\mathbf k}}

\begin{document}

\title{Fast Wavelet-Based Visual Classification}
\author{Guoshen Yu\\
\emph{CMAP, Ecole Polytechnique, 91128 Palaiseau Cedex, France}\\
\emph{yu@cmap.polytechnique.fr}\\
\and
Jean-Jacques Slotine\\
\emph{Nonlinear Systems Laboratory, Massachusetts Institute of Technology, Cambridge, MA 02139, USA}\\
\emph{jjs@mit.edu}\\
}
 \maketitle \thispagestyle{empty}

\begin{abstract}
We investigate a biologically motivated approach to fast visual
classification, directly inspired by the recent work~\cite{Serre07}.
Specifically, trading-off biological accuracy for computational
efficiency, we explore using wavelet and grouplet-like transforms to
parallel the tuning of visual cortex V1 and V2 cells, alternated
with max operations to achieve scale and translation invariance. A
feature selection procedure is applied during learning to accelerate
recognition. We introduce a simple attention-like feedback
mechanism, significantly improving recognition and robustness in
multiple-object scenes. In experiments, the proposed algorithm
achieves or exceeds state-of-the-art success rate on object
recognition, texture and satellite image classification, language
identification and sound classification.

\end{abstract}
\vspace{-5ex}
\section{Introduction}
\vspace{-2ex} Automatic object recognition and image classification
are important and challenging tasks. This paper is inspired by the
remarkable recent work of Poggio, Serre, and their
colleagues~\cite{Serre07}, on rapid object categorization using a
feedforward architecture closely modeled on the human visual system.
The main directions it departs from that work are twofold.  First,
trading-off biological accuracy for computational efficiency, our
results exploit more engineering-motivated mathematical tools such
as wavelet and grouplet
transforms~\cite{MallatWaveletTour,Mallat08}, allowing faster
computation and limiting ad-hoc parameters. Second, the approach is
generalized by adding a degree of \textit{feedback} (another known
component of human perception), yielding significant performance and
robustness improvement in multiple-object scenes.  In experiments,
the resulting scale- and translation-invariant
 algorithm achieves or exceeds state-of-the-art performance in object
recognition, but also in texture and satellite image classification,
and in language identification. \vspace{-2ex}

\section{Algorithm description}
\vspace{-2ex}
\subsection{Feature computation and classification}
\vspace{-2ex}

\begin{figure}[htbp]
\begin{center}
\vspace{-4ex}
\includegraphics[width=7cm]{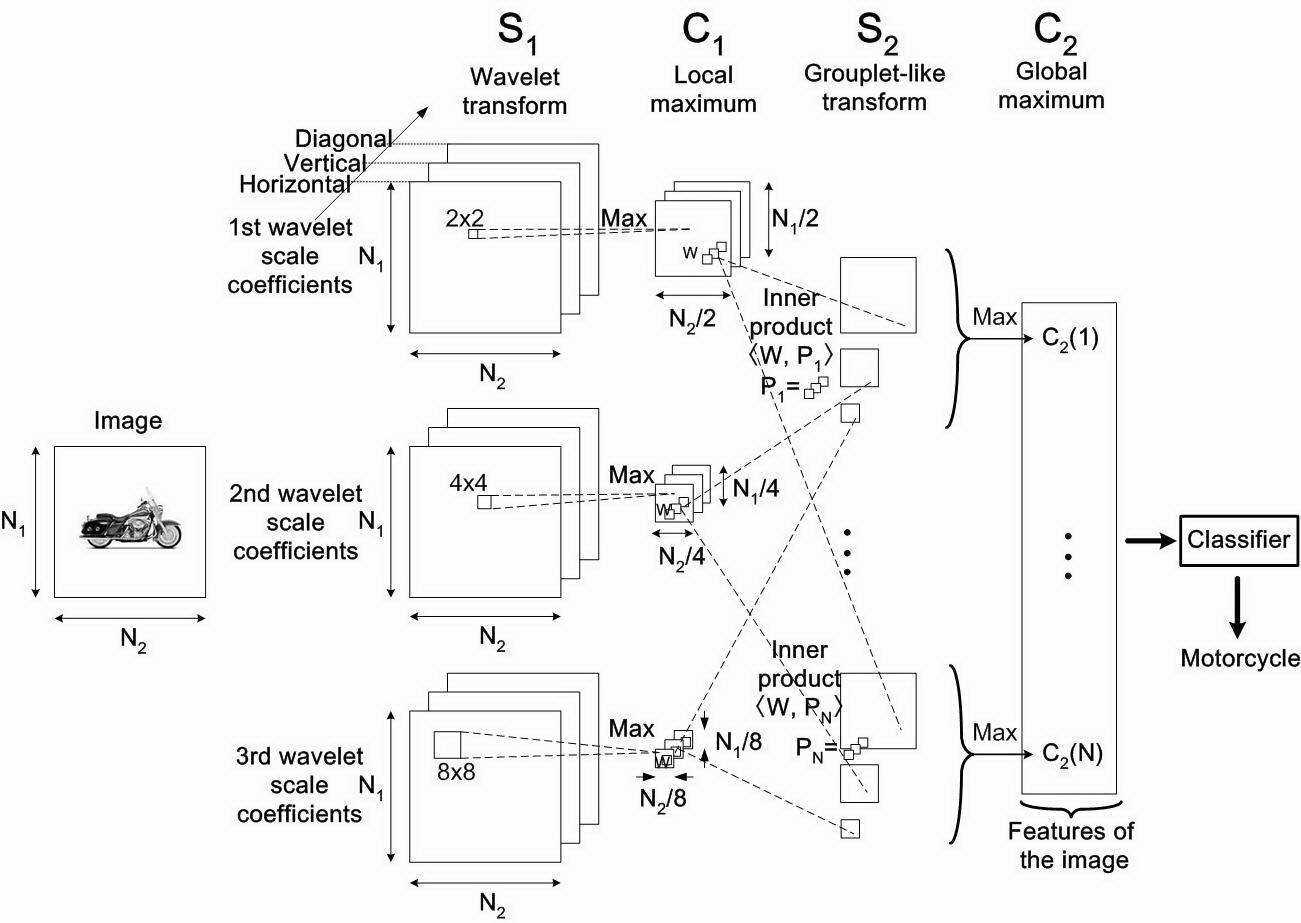}
\vspace{-5ex}
\end{center}
\caption{\small{\textbf{Algorithm overview.}}} \label{fig:algo}
\vspace{-2ex}
\end{figure}

As in~\cite{Serre07}, the algorithm is hierarchical.  In addition,
motivated in part by the relative uniformity of cortical
anatomy~\cite{Mountcastle78,Von00}, the two layers of the hierarchy
are made to be computationally similar, as shown in
Fig.~\ref{fig:algo}.  Layer one performs a wavelet
transform~\cite{MallatWaveletTour} in the $\mathbf{S_1}$ unit
followed by a local maximum operation in the $\mathbf{C_1}$ unit.
The transform in the $\mathbf{S_2}$ unit in layer two is similar to
the grouplet transform~\cite{Mallat08}, and is followed by a global
maximum
operation in the $\mathbf{C_2}$ unit.  \\

\noindent $\mathbf{S_1}:$ \textbf{Wavelet transform}. The frequency
and orientation tuning of cells in visual cortex V1 can be
interpreted as performing a wavelet transform of the retinal
image~\cite{MallatWaveletTour}. Let us denote $f(x,y)$ a gray-level
image of size $N_1 \times N_2$. A translation-invariant wavelet
transform is performed on the image: \vspace{-2ex}
\begin{equation}
 Wf(u,v,j,k) =  \sum_{x=1}^{N_1} \sum_{y=1}^{N_2}
f(x,y) \frac{1}{2^{j}} \psi^k \left( \frac{x-u, y-v}{2^{j}} \right),
\nonumber \vspace{-1ex}
\end{equation}
 where $k=1,2,3$ denotes the orientation
(horizontal, vertical, diagonal), $\psi^k(x,y)$ is a wavelet
function and $Wf$ are the wavelet coefficients. Scale invariance is
achieved by a normalization \vspace{-2ex}
\begin{equation}
\label{eqn:wavelet:norm} S_1(u,v,j,k) =
\frac{|Wf(u,v,j,k)|}{\|f\|^2_{\textrm{supp}(\psi^k_j)}}, \nonumber
\vspace{-2ex}
\end{equation}
where $\|f\|^2_{\textrm{supp}(\psi^k_j)}$ is the image energy within
the support of the wavelet $\psi^k \left( \frac{x-u, y-v}{2^{j}}
\right)$. One can verify that \vspace{-1ex}
\begin{equation}
 S_1(u,v,j,k) \sim S_1'(2^\beta u, 2^\beta v,\beta j,k) \nonumber
 \vspace{-2ex}
\end{equation}
where $S_1$ and $S_1'$ are the coefficients of $f(x,y)$ and of its
$2^\beta$-time zoomed version $f( {x}/{2^\beta}, {y}/{2^\beta} )$.
The normalization also makes the
recognition invariant to
global linear illumination change. \vspace{-2ex}\\

 \noindent $\mathbf{C_1}:$ \textbf{Local maximum}
Limited translation invariance is achieved at this stage by keeping
the local maximum of $S_1$ coefficients in a subsampling procedure:
 \vspace{-1ex}
\begin{equation}
\hspace{-4ex} C_1(u,v,j,k) = \max_{u'\in[2^{j}(u-1)+1,2^{j}u),
v'\in[2^{j}(v-1)+1,2^{j}v)} S_1(u',v',j,k), \nonumber
 \vspace{-1ex}
\end{equation}
the maximum being taken at each scale $j$ and orientation $k$ within
a spatial neighborhood of size proportional to $2^{j} \times 2^{j}$.
The resulting $C_1$ map at scale $j$ and orientation $k$ is thus of
size $N_1/2^j \times N_2/2^j$.


\noindent $\mathbf{S_2}:$ \textbf{Grouplet-like transform}. Cells in
visual cortex V2 and V4 have larger receptive fields comparing to
those in V1 and are tuned to geometrically more complex stimuli such
as contours and corners~\cite{Rao99}. The geometrical grouplets
recently proposed by Mallat~\cite{Mallat08} imitate this mechanism
by grouping and re-transforming the wavelet coefficients.

The procedure in $\mathbf{S_2}$ is similar to the grouplet
transform. Instead of grouping the wavelet coefficients with a
multi-scale geometrically adaptive association field and then
re-transforming them with Haar-like functions as in~\cite{Mallat08},
responses of $\mathbf{S_2}$ are obtained via inner products between
$C_1$ coefficients and sliding patch functions of different sizes:
\vspace{-1ex}
\begin{equation}
\hspace{-3ex} S_2(u,v,j,i) = \sum_{u'=1}^{N_1/2^j}
\sum_{v'=1}^{N_2/2^j} \sum_{k=1}^{3} C_1(u',v',j,k)
P_i(u'-u,v'-v,k), \nonumber
 \vspace{-1ex}
\end{equation}
where $P_i$ of support size $M_i \times M_i \times 3$ are patch
functions that group the 3 wavelet orientations in a square of
size $M_i \times M_i$.

While the grouplet functions are adaptively chosen to fit the
geometry in the image~\cite{Mallat08}, the patch functions $P_i$,
$i=1, \ldots,N$ are learned with a simple random sampling as
in~\cite{Serre07}: each patch is extracted at a random scale and a
random position from the $C_1$ coefficients of a randomly selected
training image, the rationale being that patterns that appear with
high probability are likely to be learned. \vspace{-2ex}
\\

\noindent $\mathbf{C_2}:$ \textbf{Global maximum}. A global maximum
operation in space and in scale is applied on $S_2$ and the
resulting $C_2$ coefficients \vspace{-1ex}
\begin{equation}
C_2(i) = \max_{u,v,j} S_2(u,v,j,i) \nonumber
 \vspace{-2ex}
\end{equation}
are thus invariant to image translation and scale change.   \vspace{-2ex} \\

\noindent \textbf{Classification} The classification uses $C_2$
coefficients as features and thus inherits the translation and scale
invariance. While various classifiers such as SVMs can be used, a
simple but robust nearest neighbor classifier will be applied in the
experiments.
 \vspace{-2ex}

\subsection{Feature selection}
 \vspace{-2ex} Structures that appear with a high probability are
likely to be learned as patch functions through random sampling.
However, they are not necessarily salient and neither are the
resulting $C_2$ features. This suggests active selection of the
learned patches.\footnote{Besides improving computational efficiency
of the algorithm, such reorganization is inspired both by a similar
process thought to occur after immediate learning, notably during
sleep, and by the relative uniformity of cortical
anatomy~\cite{Mountcastle78} which suggests enhancing computational
similarity between the two layers.} For example, Lowe and Mutch have
constructed sparse patches by retaining one salient direction at
each position~\cite{Mutch06}.

\begin{figure}[htbp]
\begin{center}
\vspace{-3ex}
\begin{tabular}{ccc}
\includegraphics[width=2cm]{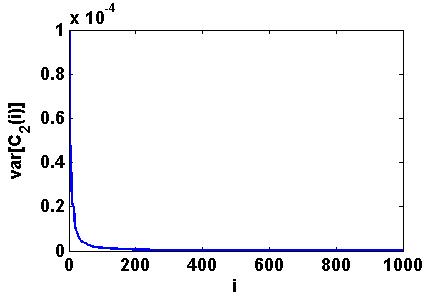}
& \hspace{-3ex}
\includegraphics[width=3cm]{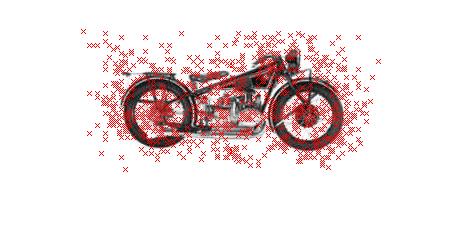} &
\hspace{-3ex}\includegraphics[width=3cm]{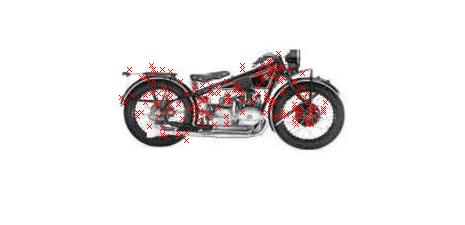} \\
\textbf{a} & \textbf{b} & \textbf{c}\\
\end{tabular}
\vspace{-6ex}
\end{center}
\caption{\small{\textbf{\textbf{a.}~Variance of the $C_2$
coefficients before patch selection. \textbf{b} and
\textbf{c.}~Positions of the $C_2$ coefficients before and after
patch selection (marked by crosses).}}} \label{fig:C2:variance}
\vspace{-4ex}
\end{figure}

A simple patch selection is proposed here by sorting the variances
of the $C_2$ coefficients of the \textit{training} images. A small
$C_2(i)$ variance implies that the corresponding patch $P_i$ is not
salient. Fig.~\ref{fig:C2:variance}-a plots the variance of the
$C_2$ coefficients of the motorcycle and the background images in
the Caltech5 database (see Fig.~\ref{fig:caltech5}), the $S_2$
patches being learned from the same images. Out of the 1000 patches,
200 salient ones whose resulting $C_2$ have non-negligible variance
are selected. Other patches usually correspond to nonsalient
structures such as a common background and are therefore excluded.
Fig.~\ref{fig:C2:variance}-b and c show that after patch selection
the 200 $C_2$ coefficients are mainly positioned around the object,
as opposed to the 1000 $C_2$ coefficients spreading over all the
image prior to patch selection. The recognition using these salient
patches is not only more robust but also 5 times faster.

\vspace{-3ex}
\subsection{Feedback}
\vspace{-2ex} \label{subsec:feedback}
Feedback~\cite{Rao99,Hawkins04,Walther05} allows tracing back object
positions, focusing attention on the objects one by one and thus
improving recognition performance in multiple-object scenes. \vspace{-2ex}\\

\begin{figure}[htbp]
\begin{center}
\vspace{-2ex}
\begin{tabular}{ccc}
\includegraphics[width=2cm]{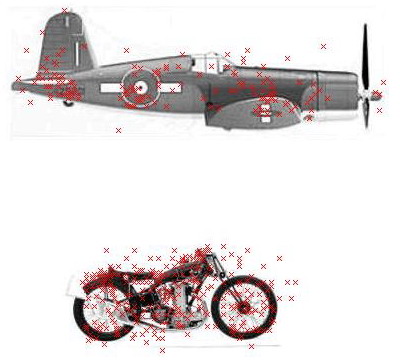}
&
\includegraphics[width=2cm]{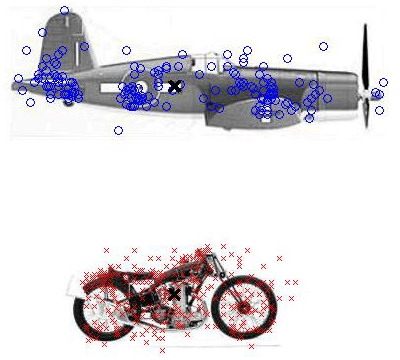}
&
\includegraphics[width=3cm]{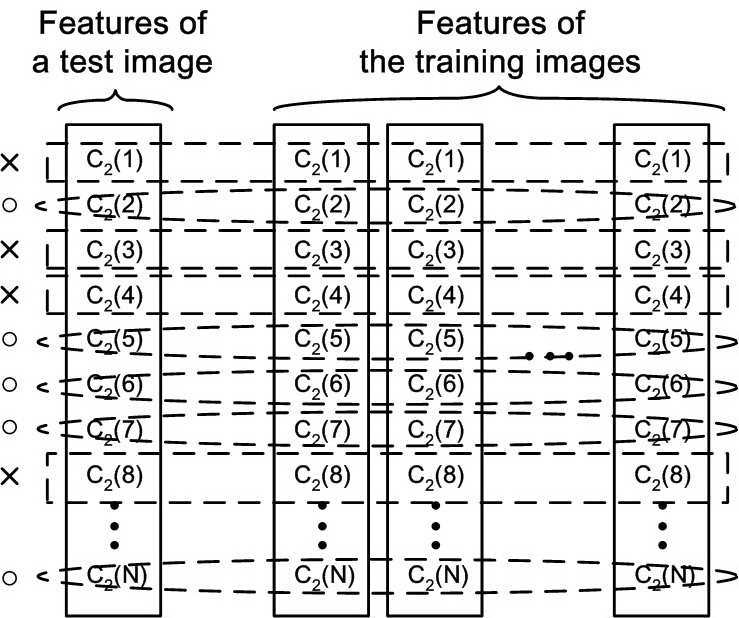} \\
\textbf{a} & \textbf{b} & \textbf{c}
\\
\vspace{-6ex}
\end{tabular}
\end{center}
\caption{\small{\textbf{Feedback in a two-object scene.
\textbf{a.}~Positions of $C_2$ coefficients are marked by crosses.
\textbf{b.}~$C_2$ coefficients are clustered (represented by circles
vs crosses). \textbf{c.} Feature coefficients of the
\textit{training} images are grouped, the coordinates being in line
with the clustering of the coefficients of the \textit{test} image.
Rectangles and ellipses represent the two groups.}}}
\label{fig:C2:feedback:2obj} \vspace{-5ex}
\end{figure}

\noindent \textbf{Object positioning} For simplicity the feedback
procedure is discussed in a two-object scene but can be applied in
the case of multiple objects. $C_2$ coefficients are placed around
the two objects after selection, as shown in
Fig.~\ref{fig:C2:feedback:2obj}-a. Using a clustering algorithm such
as the K-means algorithm, one is able to locate the two objects as
illustrated in Fig.~\ref{fig:C2:feedback:2obj}-b. \vspace{-2ex} \\

\noindent \textbf{Object identification} While one could recalculate
the features of the attended object cropped out from the whole
image, i.e., concentrate all the visual cortex resource on a single
object, a faster procedure identifies the attended object, say
object $A$, using directly the lower-dimensional feature vector
$C_{2A}$, composed of the $C_2$ coefficients corresponding to $A$
already calculated in the feedforward pathway. This can be
implemented by reclassifying $C_{2A}$ using subsets of the $C_2$
coefficients of the \textit{training} images extracted at the same
coordinates of $C_{2A}$, as shown in
Fig.~\ref{fig:C2:feedback:2obj}-c. Discarding the coordinates that
are located on the irrelevant object $B$ in the test image
disambiguates the classification and improves the recognition of the
object $A$.

\vspace{-2ex}
\section{Experiments}
\vspace{-2ex} All the experiment results were obtained with the same
algorithm configuration. Daubechies 7-9 wavelets of 3 scales were
used in $\mathbf{S_1}$. In $\mathbf{S_2}$ 1000 patches $P_i$ of 4
different sizes $M \times M \times 3$ with $M=4,8,12,16$, 250 for
each, were learned from the training images. The classifier was the
simple nearest neighbor classification algorithm.

For texture and satellite image classification as well as for
language identification, one sample image of size $512 \times 512$
was available per image class and was segmented to 16
non-overlapping parts of size $128 \times 128$. Half were used for
training and the rest for test.

\vspace{-2ex}
\subsection{Object recognition}
\begin{figure}[htbp]
\begin{center}
\vspace{-2ex}
\begin{tabular}{ccccc}
\includegraphics[width=1cm]{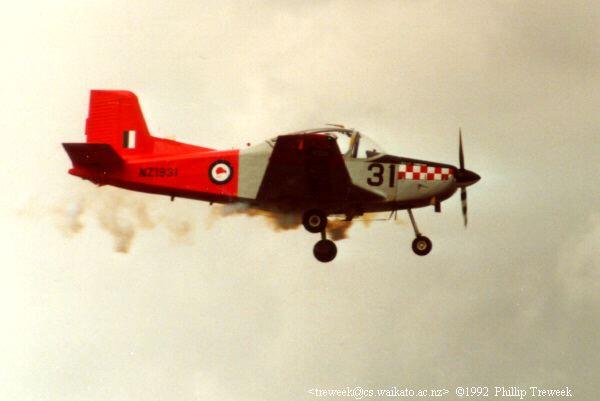}
&
\includegraphics[width=1cm]{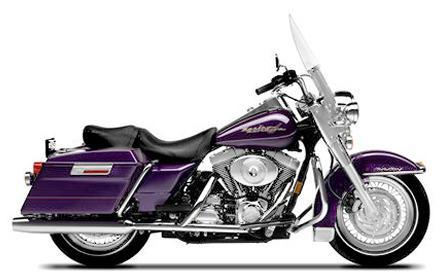}
&
\includegraphics[width=1cm]{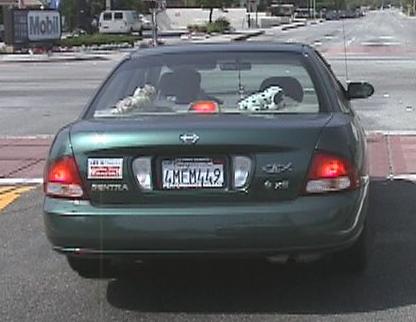}
&
\includegraphics[width=1cm]{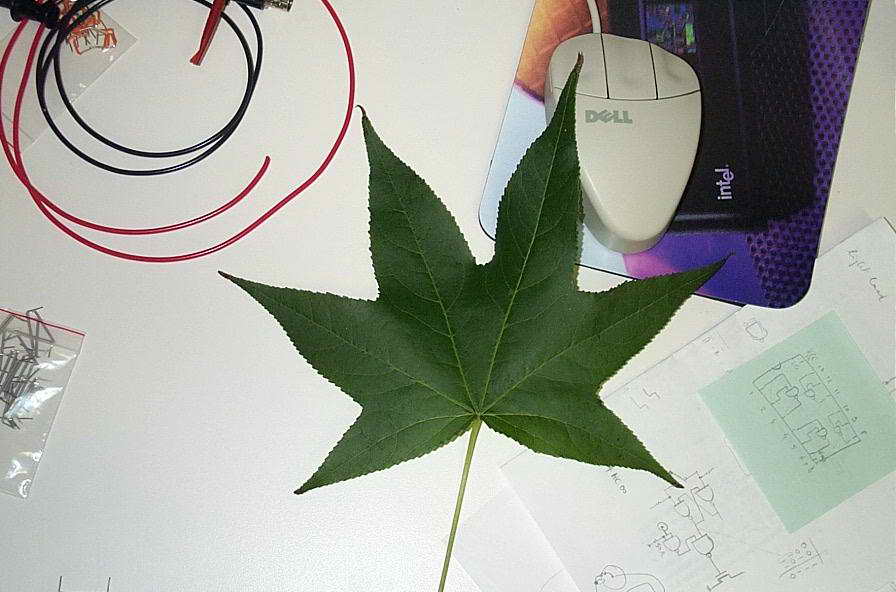}
&
\includegraphics[width=1cm]{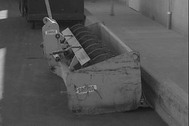}
\\
\includegraphics[width=1cm]{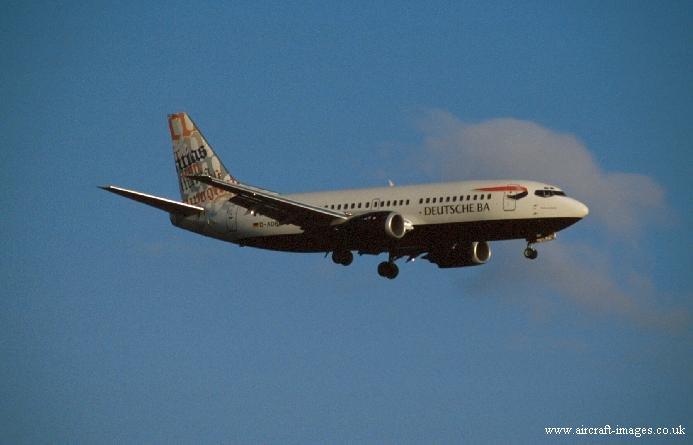}
&
\includegraphics[width=1cm]{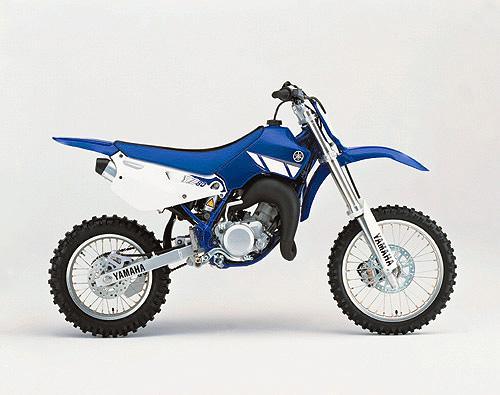}
&
\includegraphics[width=1cm]{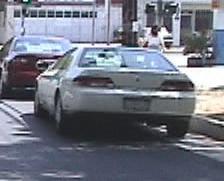}
&
\includegraphics[width=1cm]{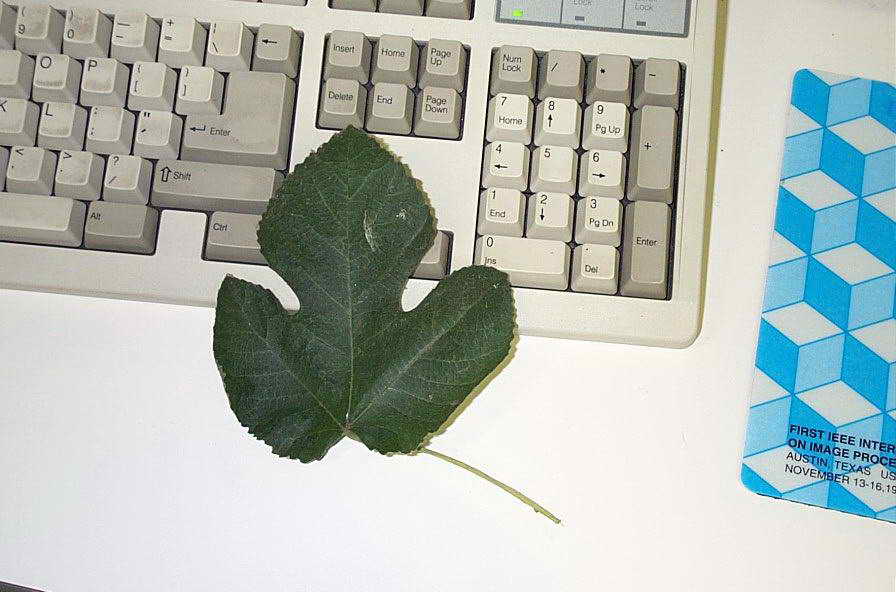}
&
\includegraphics[width=1cm]{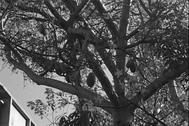}
\\
\vspace{-6.5ex}
\end{tabular}
\end{center}
\caption{\small{\textbf{Sample images from Caltech5. From left to
right: airplanes, motorcycles, cars (rear), leaves and
background.}}}
 \label{fig:caltech5} \vspace{-6ex}
\end{figure}

\vspace{-2ex}
 For the object recognition experiments we used 4
data sets that are airplanes, motorcycles, cars (rear) and leaves,
plus a background class from the Caltech5
database\footnote{\scriptsize{http://www.robots.ox.ac.uk/$\sim$vgg/data3.html}},
some sample images being shown in Fig.~\ref{fig:caltech5}. The
images are turned to gray-level and rescaled in preserving the
aspect ratio so that the minimum side length is of 140 pixels. A set
of 50 positive images and 50 negative images were used for training
and another set for test.

\begin{figure}[htbp]
\begin{center}
\vspace{-1ex}
\begin{tabular}{cc}
\includegraphics[width=2.5cm]{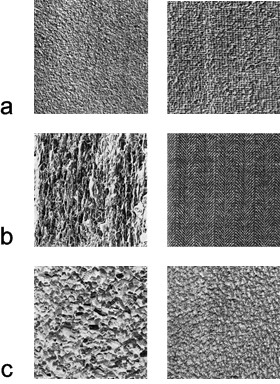}
& \hspace{-2ex}
\includegraphics[width=6cm]{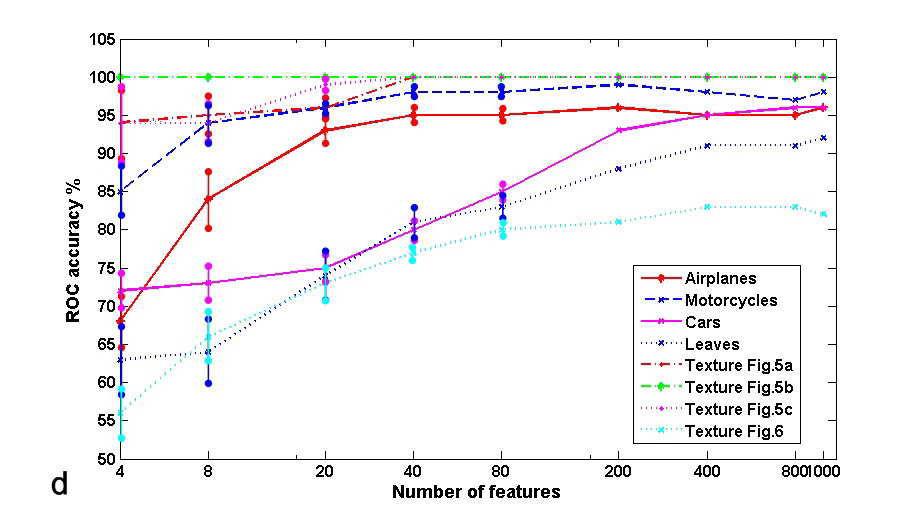}
\\
\vspace{-7ex}
\end{tabular}
\end{center}
\caption{\small{\textbf{a-c.~Pairs of textures. d.~Performance vs
numbers of $C_2$ features.}}}
 \label{fig:accuracy:vs:num:features} \vspace{-4ex}
\end{figure}

Table~\ref{tab:object:binary} summarizes the object recognition. The
performance measure reported is the ROC
accuracy.\footnote{\scriptsize{ROC accuracy: $R=1-((1-p)x+p(1-y))$,
where $x$ and $y$ are respectively the false positive rate on the
negative samples and true positive rate on the positive samples, $p$
is the proportion of the positive samples.}} Results obtained with
the proposed algorithm are superior to previous
approaches~\cite{Fergus03, Weber00} and comparable to~\cite{Serre07}
but at a lower computational cost (in Matlab code about 6 times
faster with feature selection).
Fig.~\ref{fig:accuracy:vs:num:features}-d shows that the performance
is improved when the number of $C_2$ features increases and is in
general stable with 200 features.

\begin{table}[htb]
\begin{center}
\begin{tabular}{|c|c|c|c|}
\hline Data sets & Proposed & Serre~\cite{Serre07} & Others \\
\hline
Airplanes & 96.0 & 96.7 & 94.0~\cite{Fergus03}\\
\hline
Motorcycles & 98.0 & 98.0 & 95.0~\cite{Fergus03} \\
\hline
Cars (Rear) & 96.0 & 99.8 & 84.8~\cite{Fergus03} \\
\hline
Leaves & 92.0 & 97.0 & 84.0~\cite{Weber00} \\
 \hline
\end{tabular}
\vspace{-3ex}
\end{center}
\caption{\small{\textbf{Object recognition performance.}}}
\label{tab:object:binary} \vspace{-3.5ex}
\end{table}

\vspace{-2ex}
\subsection{Texture classification}
\begin{figure}[htbp]
\begin{center}
\begin{tabular}{ccccc}
\includegraphics[width=1cm]{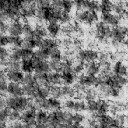}
&
\includegraphics[width=1cm]{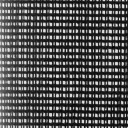}
&
\includegraphics[width=1cm]{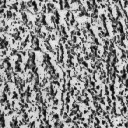}
&
\includegraphics[width=1cm]{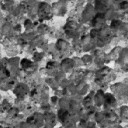}
&
\includegraphics[width=1cm]{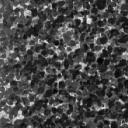}
\\
\includegraphics[width=1cm]{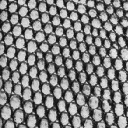}
&
\includegraphics[width=1cm]{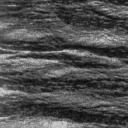}
&
\includegraphics[width=1cm]{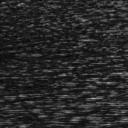}
&
\includegraphics[width=1cm]{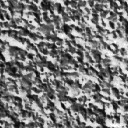}
&
\includegraphics[width=1cm]{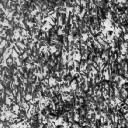}
\\
\vspace{-6ex}
\end{tabular}
\end{center}
\caption{\small{\textbf{A group of 10 textures.}}}
\label{fig:texture:10classes} \vspace{0ex}
\end{figure}

\vspace{-2ex} Figs.~\ref{fig:accuracy:vs:num:features}-a,b,c and
Fig.~\ref{fig:texture:10classes} show respectively 3 pairs of
textures that were used for binary classification and a group of 10
textures that were used for multiple-class (10-class)
classification, all from the Brodatz
database\footnote{Fig.~\ref{fig:accuracy:vs:num:features}-a,b,c: D4
and D84; D12 and D17; D5 and D92. Fig.~\ref{fig:texture:10classes}
 : D4, D9, D19, D21, D24, D28, D29, D36, D37, D38.}. As summarized in
Table~\ref{tab:texture}, the proposed algorithm achieved perfect
results for binary classification and for the challenging multiple
class classification its performance was comparable to the
state-of-the-art methods~\cite{Liu03,kim02,Randen99}. Indeed the
random patch extraction applied in the algorithm is ideal for
classifying stationary patterns such as textures.
Fig.~\ref{fig:accuracy:vs:num:features} shows that stable
performance is achieved with as few as 40 features, which confirms
the good texture classification results and the robustness of the
algorithm.


\begin{table}[htb]
\begin{center}
\vspace{-2ex}
\begin{tabular}{|c|c|c|c|c|}
\hline Data sets & Proposed & Avg.~\cite{Randen99}  & Best~\cite{Randen99} & Others \\
\hline
Fig.~\ref{fig:accuracy:vs:num:features}-a & 100 & 88.0 & 99.3 & 91.5~\cite{kim02} \\
\hline
Fig.~\ref{fig:accuracy:vs:num:features}-b & 100 & 96.2 & 99.7 & N/A \\
\hline
Fig.~\ref{fig:accuracy:vs:num:features}-c & 100 & 87.6 & 97.5 & 88.6~\cite{kim02} \\
\hline
Fig.~\ref{fig:texture:10classes} & 82.6 & 52.6 & 67.7 & 83.1~\cite{Liu03} \\
 \hline
\end{tabular}
\vspace{-4ex}
\end{center}
\caption{\small{\textbf{Texture classification performance. From
left to right: proposed method, average and best performance of the
algorithms summarized in~\cite{Randen99}, other methods. N/A means
that the results were not shown.}}}
 \label{tab:texture}
\end{table}

Classifying the whole Brodatz database (111 textures) is a more
challenging task. Combining $C_2$ coefficients with the histogram of
the wavelet approximation coefficients as features, the proposed
algorithm achieved 87.8\% accuracy for the 111-texture
classification, comparable to the 88.2\% accuracy rate reported
in~\cite{Lazebnik05} obtained with a state-of-the-art texture
classification approach.

\subsection{Satellite image classification}
\begin{figure}[htbp]
\begin{center}
\vspace{-2ex}
\begin{tabular}{cccc}
\includegraphics[width=1.5cm]{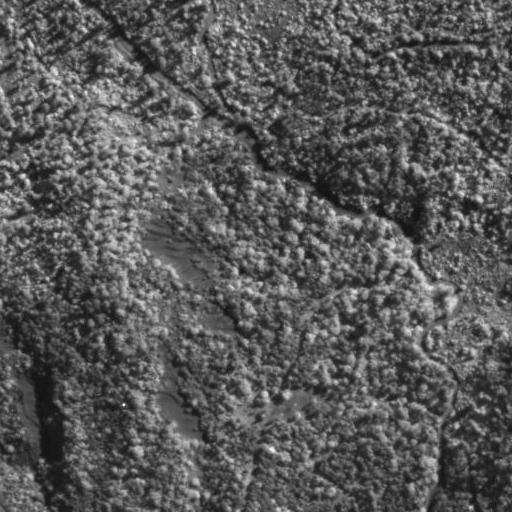} &
\includegraphics[width=1.5cm]{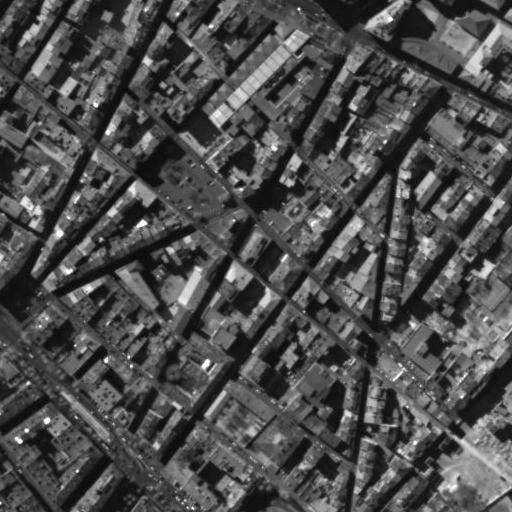} &
\includegraphics[width=1.5cm]{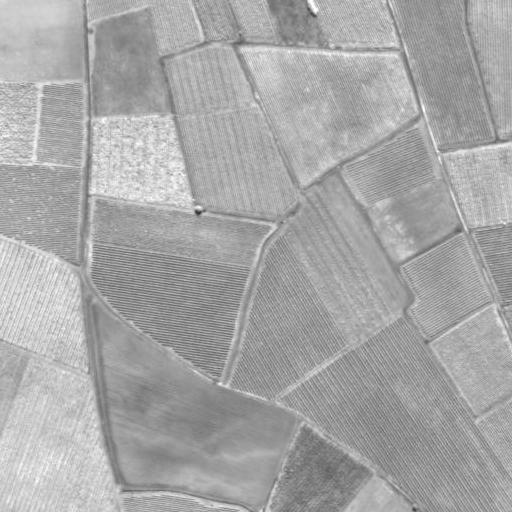} &
\includegraphics[width=1.5cm]{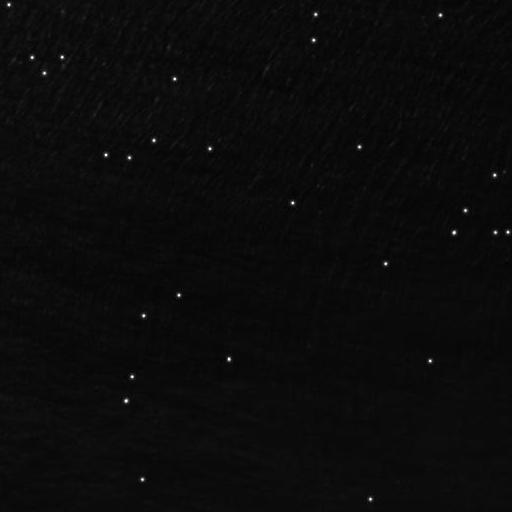} \\
\vspace{-7ex}
\end{tabular}
\end{center}
\caption{\small{\textbf{Satellite images. From left to right:
forest, urban areas, rural areas, sea.}}} \label{fig:satellite}
\vspace{-3ex}
\end{figure}

Fig.~\ref{fig:satellite} displays 4 classes of satellite images at
0.5 $m$ resolution: urban areas, rural areas, forests and sea. Since
access to images at other resolutions is restricted, we simulated
the images at resolutions 1 $m$ and 2 $m$ by Gaussian convolution
and sub-sampling.

The first experiment tested the multi-class classification of
mono-resolution images shown in Fig.~\ref{fig:satellite}. 100\%
classification accuracy was achieved for images of all the 4
classes. The second experiment validated the scale invariance of the
proposed algorithm. Images at resolution 0.5 $m$ were used to train
the classifier while the classification was tested on images at
resolutions 1 $m$ and 2 $m$. Again the classification accuracy was
100\%, same as reported in a recent work~\cite{Luo08} and
significantly higher than earlier methods~\cite{Haralick73}
referenced therein. In addition, image resolution is assumed to be
known in~\cite{Luo08}, whereas the proposed algorithm does not need
this information, thanks to its scale invariance.

\subsection{Language identification}
\begin{figure}[htbp]
\begin{center}
\vspace{-1ex}
\begin{tabular}{cccc}
\hspace{-1ex}\includegraphics[width=1.5cm]{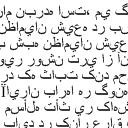} &
\hspace{-1ex}\includegraphics[width=1.5cm]{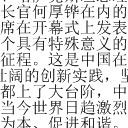} &
\hspace{-1ex}\includegraphics[width=1.5cm]{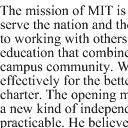} &
\hspace{-1ex}\includegraphics[width=1.5cm]{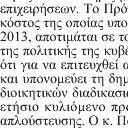}
\\
\hspace{-1ex}\includegraphics[width=1.5cm]{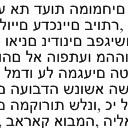} &
\hspace{-1ex}\includegraphics[width=1.5cm]{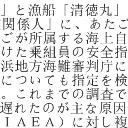} &
\hspace{-1ex}\includegraphics[width=1.5cm]{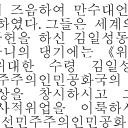} &
\hspace{-2ex} \includegraphics[width=1.5cm]{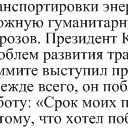} \\
\end{tabular}
\end{center}
\caption{\small{\textbf{From top to bottom, left to right: document
texts in Arabic, Chinese, English, Greek, Hebrew, Japanese, Korean,
Russian.}}} \label{fig:languages}
\end{figure}

Language identification aims to determine the underlying language of
a document in an imaged format, and is often carried out as a
 preprocessing of optical character recognition (OCR).  Based on
 principles totally different from traditional approaches~\cite{Lu08},
 the proposed algorithm achieved 100\% success rate in a 8-language
 identification task, as shown in Fig~\ref{fig:languages}.

\subsection{Sound Classification}
The main idea is to directly extend the above algorithm to sound
applications is to view time-frequency representations of sound as
textures. Preliminary experiments suggest this may be a fruitful
direction of research.

\begin{figure}[htbp]
\begin{center}
\begin{tabular}{ccccc}
\hspace{-4ex}{\includegraphics[width=1.5cm]{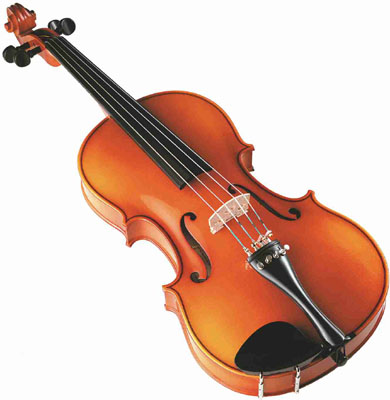}} &
\hspace{-1ex}{\includegraphics[width=1.5cm]{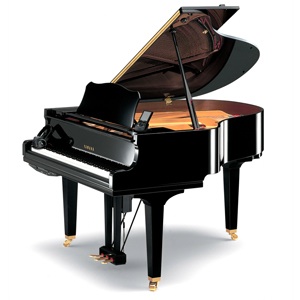}} &
\hspace{-1ex}{\includegraphics[width=1.5cm]{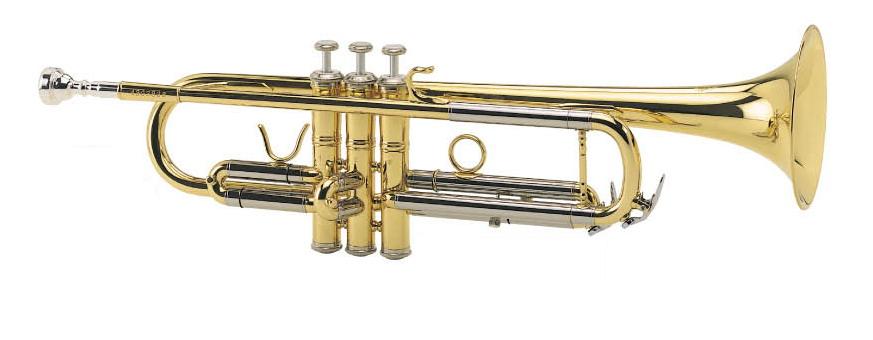}}
& \hspace{-1ex}{\includegraphics[width=1.5cm]{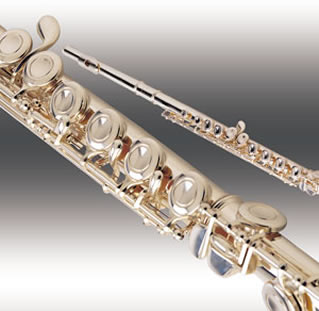}}
& \hspace{-1ex}{\includegraphics[width=1.5cm]{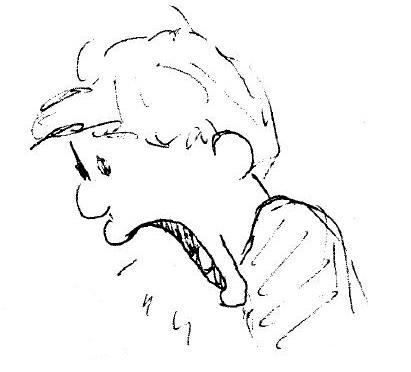}}
\vspace{3ex}\\
\hspace{-4ex}\includegraphics[width=1.5cm]{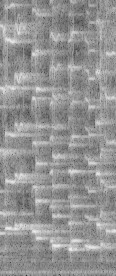}
&
\hspace{-1ex}\includegraphics[width=1.5cm]{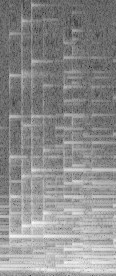}
&\hspace{-1ex}\includegraphics[width=1.5cm]{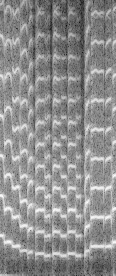}
&
\hspace{-1ex}\includegraphics[width=1.5cm]{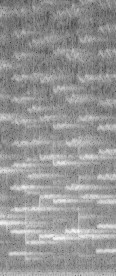}
& \hspace{-1ex}\includegraphics[width=1.5cm]{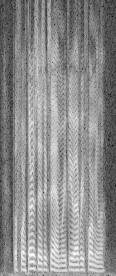} \\
\end{tabular}
\end{center}
\caption{\small{\textbf{From left to right: violin, piano, trumpet,
flute, speech. The figures in the second row are spectrograms of the
sounds illustrated in the first row.}}} \label{fig:sounds}
\end{figure}

Fig.~\ref{fig:sounds} illustrates 5 types of sounds and samples of
their log-spectrograms. 2 minutes excerpts of each sound were
collected. The spectrograms were segmented (in time) into segments
of 5 seconds. Half were used for training and the rest for test. A
direct application of the proposed algorithm using the spectrograms
as the visual patterns resulted in 100\% accuracy in the 5-sound
classification.

\subsection{Feedback: multiple-object scenes}
\begin{figure}[htbp]
\begin{center}
\vspace{-4ex}
\begin{tabular}{c}
\includegraphics[width=6cm]{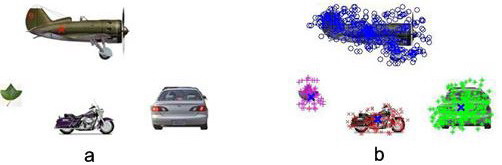}
\end{tabular}
\vspace{-5ex}
\end{center}
\caption{\small{\textbf{\textbf{a.}~A 4-object scene. \textbf{b.}
$C_2$ coefficients clustering.}}} \label{fig:multiple:object}
\end{figure}

Recognition performance tends to degrade when multiple stimuli are
presented in the receptive field. Fig.~\ref{fig:multiple:object}-a
shows an example of a multiple-object scene in which one searched an
object, say an airplane, through a binary classification against a
background image. Due to the perturbation from the coexisting
stimuli, the feedforward recognition accuracy is as low as 74\%. The
feedback procedure introduced in Subsection~\ref{subsec:feedback}
improves considerably the accuracy to 98\% by focusing attention on
each object in turn. \vspace{-0ex}

\section{Conclusion and future work}
\vspace{-3ex} Inspired by the biologically motivated work
of~\cite{Serre07}, we have described a wavelet-based algorithm which
can compete with the state-of-the-art methods for fast and robust
object recognition, texture and satellite image classification,
language recognition and sound classification. A feedback procedure
has been introduced to improve recognition performance in
multiple-object scenes.

Potential applications also include video archiving (semantic video
analysis), video surveillance,  high-throughput drug development,
texture retrieval,  and robotic learning by imitation.

To further improve and extend the algorithm, a key aspect will be a
more refined use of feedback between different levels. Such feedback
will naturally involve stability and convergence questions, which
will in turn both guide the design of the algorithm and shape its
performance.  In addition, contrary to the nervous system, the
algorithm need not be constrained by information transmission delays
between different levels.  Preliminary ideas in this direction are
briefly discussed in the appendix. \\

\small{\noindent \textbf{Acknowledgements:} We are grateful to
Tomaso Poggio and Thomas Serre for many discussions about their
recognition system, to St\'ephane Mallat for stimulating discussions
on wavelets and grouplets and to Jean-Michel Morel for important
discussions on invariant image recognition.}

\appendix

\begin{center}
   {\bf APPENDIX}
\end{center}

\section{Dynamic System Perspective}
\subsubsection{Basic algorithm}
The first step towards introducing a dynamic systems perspective
aimed at further development of feedback mechanisms is simply to
rewrite the algorithm in terms of differential equations, which puts
it in a form more suitable to subsequent analysis of stability and
convergence.

Let ${\bx_1}$ be the output of the $S_1/C_1$ layer, and ${\bx_2}$ be
the output of the $S_2/C_2$ layer (in our static implementation
above, we simply have ${\bx_1} = C_1$ and ${\bx_2} = C_2$).

For a single object, the basic algorithm can be trivially computed
by a dynamic system of the form

$\dot{\bx}_1 = - k_1 ({\bx}_1 - C_1)$

$\dot{\bx}_2 = - k_2 ({\bx}_2 - C_2)$

For multiple objects, the clustering process described in
section~\ref{subsec:feedback} can be implemented by introducing a
scalar state ${x_3}$, which spikes for each object in sequence

$\dot{x}_3 = \rho(x_3,C_2)$

\noindent with spike amplitude equal to $1$ $-$ the function
$\rho(x_3,C_2)$ is discussed later in this section and in a
companion paper. The dynamics of ${\bx}_2$ can be modified in turn
so that states corresponding to each object appear in sequence
according to the state $x_3$

$\dot{\bx}_1 = - k_1 ({\bx}_1 - C_1)$

$\dot{\bx}_2 = - k_2 ({\bx}_2 - x_3 \ \bk(C_{2A}))$

$\dot{x}_3 = \rho(x_3,C_2) $

\noindent where, componentwise, $\bk(C_{2A}) = C_{2A}$ where
$C_{2A}$ is active and and $\bk(C_{2A}) = 0$ otherwise. Note that
$x_3$ smoothly transitions between $0$ and $1$ according to the
attended object. The positive gain $k_2$ is chosen such that $k_2 T
\gg 1$, where $T$ is the spike duration, itself a fraction of the
interspike period.

The above equations simply implement the basic algorithm and display
objects in sequence, without introducing any new feature at this
point.

Techniques for globally stable spike-based clustering are described
in a companion paper, based on modified FitzHugh-Nagumo neural
oscillators~\cite{FitzHugh61,Nagumo62}, similar to~\cite{Chen02},
\begin{eqnarray}\label{FN_sat}
\label{eqn:oscillator:FN} \dot{v}_i & = & 3 v_i - v_i^3 - v_i^7 + 2
- w_i + I_i \\
\dot{w}_i & = & c [\alpha (1+\tanh(\beta v_i)) - w_i]\nonumber
\end{eqnarray}
where $v_i$ is the membrane potential of the oscillator, $w_i$ is an
internal state variable representing gate voltage, $I_i$ represents
the external current input, and $\alpha$, $\beta$ and $c$ are
strictly positive constants. Using a diagonal metric transformation
${\mathbf \Theta} = {\rm diag}(\sqrt{c \alpha \beta},1)$, one easily
shows, similarly to~\cite{wang05a}, that

$$ {\mathbf \Theta} {\mathbf J} {\mathbf \Theta}^{-1} \ < \ {\rm
diag}( \ 3 + \frac{\alpha \beta}{4}\ ,\ 0\ ) $$

\noindent where ${\mathbf J}$ is the Jacobian matrix of
$(\ref{FN_sat})$, leading to simple global stability conditions
based on~\cite{Pham07} (section 2.2).

\subsection{Generalized diffusive connections}

   One of the most immediate additional feedback mechanisms to be
explored is that of generalized diffusive connections
(\cite{Pham07}, section 3.1.2). In a feedback hierarchy, these
correspond to achieving consensus between multiple processes of
different dimensions.

\subsection{Tracking of time-varying images}

Similarly to~\cite{Lohmiller99}, composite variables for dynamic
tracking can be used at every level, based on both top-down an
bottom-up information. This allows one to implicitly introduce
time-derivatives of signals in the differential equations, without
having to measure or compute these terms explicitly.

\vspace{-2ex}
\bibliographystyle{latex8}

\end{document}